**Multimodal Ordinal Modeling of Alzheimer's Disease Severity Using Structural MRI and Clinical Data**

Boris-Stephan Rauchmann[1,2*], Jonathan Laib[2], Buse Ercik[2], Robert Perneczky[2], Sergio Altares-López[1]

[1] Department of Neuroradiology, LMU University Hospital, Ludwig Maximilian University of Munich, Munich, Germany

[2] Department of Psychiatry and Psychotherapy, LMU University Hospital, Ludwig Maximilian University of Munich, Munich, Germany

*Corresponding Author: boris.rauchmann@med.uni-muenchen.de

**Abstract:**
Neurodegenerative diseases such as Alzheimer's disease (AD) require accurate and scalable tools for assessing disease severity, yet current clinical staging remains time-intensive and prone to variability. We propose an attention-enhanced multimodal machine learning framework with ordinal regression for automated and interpretable AD severity staging. The framework integrates T1-weighted MRI with demographic and genetic variables and compares unimodal and multimodal architectures using ordinal and non-ordinal prediction heads. Models were trained and validated using cohort-stratified splits derived from the ADNI, AIBL, and NIFD datasets. A strictly held-out test set was constructed using subjects excluded from all training, validation, preprocessing, and hyperparameter tuning procedures, with subject-level splitting employed throughout to prevent data leakage. Among unimodal approaches, the T1-weighted MRI model achieved slightly higher adjacent-stage accuracy (0.963) and agreement with clinical staging (QWK 0.444) than the tabular model (QWK 0.433). Integrating imaging, demographic, and genetic information improved overall performance. The multimodal non-ordinal baseline achieved the lowest prediction error (MAE 0.340), whereas the ordinal multimodal model achieved the highest adjacent-stage accuracy (0.970) and strongest agreement with clinical staging (QWK 0.549). These findings indicate that ordinal formulations better capture the ordered structure of the CDR scale and yield predictions more consistent with clinical staging. Explainability analyses using Grad-CAM++ and SHAP demonstrated anatomically and clinically plausible model behavior, supporting transparent decision-making. Overall, attention-based multimodal learning with ordinal regression represents a robust, interpretable, and scalable approach for automated AD severity staging and AI-assisted clinical decision support.

## 1. Introduction

Neurodegenerative diseases represent a growing public health challenge in aging societies, with Alzheimer's disease (AD) being the most prevalent cause of dementia worldwide. AD is characterized by a progressive decline in cognitive and functional abilities, driven by widespread neurodegeneration and synaptic dysfunction. Neuropathologically, the disease is associated with the accumulation of amyloid-β plaques and neurofibrillary tangles composed of hyperphosphorylated tau protein, ultimately leading to neuronal loss and cortical atrophy. These pathological changes begin many years before the onset of clinical symptoms, giving rise to a prolonged preclinical and prodromal phase during which early detection and accurate staging are of paramount importance [17]. This is particularly relevant in light of the recent development and increasing availability of disease-modifying therapies that are most effective when administered early in the disease course, within a relatively narrow therapeutic window [21].

In clinical practice, disease severity in AD is commonly assessed using standardized cognitive and functional rating scales. Among these, the Clinical Dementia Rating (CDR) scale is widely regarded as a gold standard for staging dementia severity [10]. The CDR evaluates six cognitive and functional domains through semi-structured interviews with patients and informants, yielding a global score ranging from no impairment (CDR 0) to severe dementia (CDR 3). While clinically well validated, CDR assessment is time-intensive, requires trained raters, and is subject to inter-rater variability [14]. These limitations motivate the development of automated multimodal and objective tools that can support clinicians by providing consistent and scalable estimates of disease severity.
Neuroimaging biomarkers play a central role in the objective assessment of neurodegeneration. Structural magnetic resonance imaging (MRI), and in particular T1-weighted MRI, provides high-resolution information about brain anatomy and enables the quantification of regional atrophy patterns associated with AD progression [9, 4]. T1-weighted MRI is routinely acquired in both clinical and research settings and has been shown to capture characteristic volume loss in medial temporal, parietal, and frontal regions that correlate with cognitive decline and CDR stage [19, 20]. In addition, demographic and genetic risk factors such as age, sex, education level, and apolipoprotein E (APOE) genotype substantially influence the risk and trajectory of AD [15]. Integrating such tabular information with neuroimaging data has the potential to improve staging accuracy by capturing complementary aspects of disease vulnerability and progression that are not directly observable from imaging alone.

However, effectively combining heterogeneous data modalities remains a non-trivial methodological challenge. Deep learning approaches, particularly convolutional neural networks (CNNs), have demonstrated strong performance in neuroimaging-based diagnosis and prognosis of AD [11, 1]. Recent architectures leveraging residual connections and transfer learning have enabled the extraction of increasingly sophisticated anatomical representations from structural MRI data [7, 3]. Nevertheless, the majority of existing deep learning studies formulate disease staging as a standard multi-class classification problem, implicitly treating CDR stages as nominal categories. This ignores the inherently ordered nature of clinical severity and may result in implausible predictions, such as assigning higher probability to distant disease stages than to adjacent ones [6].

Ordinal regression provides a principled alternative for modeling disease severity by explicitly incorporating the ordering of stages into the learning objective [5, 12]. Recent developments such as CORAL (COnsistent RAnk Logits) and CORN (Conditional Ordinal Regression for Neural networks) enable rank-consistent predictions within deep neural networks, ensuring monotonic probability distributions across disease stages [2, 18]. Despite their clear relevance for clinical staging tasks, these ordinal frameworks remain underutilized in deep learning approaches for AD severity assessment, particularly in combination with multimodal data.

Another critical requirement for the deployment of artificial intelligence systems in clinical settings is interpretability. Clinicians need to assess whether model predictions are driven by anatomically and clinically meaningful patterns rather than spurious correlations. Explainable AI (XAI) techniques such as gradient-based saliency methods for neuroimaging data and feature attribution methods for tabular variables provide essential tools for validating model behavior and building clinical trust [16, 13, 8].

In this work, we propose an attention-enhanced multimodal deep learning framework for automated staging of clinical severity according to the CDR scale, based on T1-weighted MRI and demographic and genetic tabular data. The proposed approach integrates modality-specific feature extractors with attention-based fusion mechanisms and modern ordinal regression heads to produce rank-consistent predictions of disease severity. Using data from multiple large-scale neuroimaging cohorts, we systematically compare unimodal and multimodal architectures and evaluate the impact of ordinal modeling on staging accuracy, clinical coherence, and interpretability.

The main contributions of this study are:

(i) Developing an attention-based multimodal fusion architecture that integrates structural MRI with demographic and genetic risk factors;
(ii) Application of state-of-the-art ordinal regression frameworks to automated CDR staging, explicitly modeling the progressive nature of cognitive decline;
(iii) Provide comprehensive interpretability analyses that link model predictions to anatomically and clinically plausible features.

## 2. Materials and Methods

### *2.1. Dataset description*

This study investigates automated AD staging according to the Clinical Dementia Rating (CDR) scale by integrating structural MRI with demographic and genetic tabular data in a retrospective, multi-cohort analysis. Data were drawn from three large publicly available studies - ADNI [22], AIBL [23], and NIFD [24] - providing both a sufficiently large sample and heterogeneity in scanner types, acquisition protocols, and participant demographics to realistically evaluate model robustness and generalization. Disease severity was assessed using the global CDR score, and due to the very limited number of severe dementia cases (CDR 2 and 3), only stages 0, 0.5, and 1 were included. These stages were encoded as ordered labels to preserve the natural progression of clinical severity, reflecting relevance and allowing modeling of disease severity as an ordinal variable.

Demographic and genetic characteristics stratified by CDR stage are summarized in Table 1, while the distribution of CDR stages across the different cohorts is shown in Table 2. Statistical comparisons across CDR stages were performed to identify significant differences: continuous variables (age, years of education) were compared using one-way ANOVA, while categorical variables (sex, APOE ε4 carrier status) were compared using chi-square tests. Statistical comparisons indicated significant differences across CDR stages ($p < 0.001$ for all variables), although effect sizes for some variables (e.g., age and education) were small, reflecting the large sample size. Missing values were primarily observed for years of education (24.3%), while other variables showed minimal missingness (<6%). Missing tabular values were imputed using multivariate imputation by chained equations (MICE) restricted to the training set to prevent information leakage.

**Table 1.** Demographic and genetic characteristics of the study population stratified by CDR stage.

| Variable | CDR 0 | CDR 0.5 | CDR 1 | Total | p-value |
|---|---|---|---|---|---|
| **N** | 3001 | 2276 | 699 | 5976 | - |
| **Age (years) mean ± SD** | 73.5 ± 6.9 | 74.7 ± 7.3 | 75.3 ± 7.8 | 74.1 ± 7.2 | <0.001 |
| **Education (years) mean ± SD** | 16.6 ± 2.6 | 15.8 ± 2.8 | 15.4 ± 3.0 | 16.2 ± 2.8 | <0.001 |
| **Female (%)** | 53.2 | 41.8 | 45.5 | 47.9 | <0.001 |
| **APOE ε4 carriers (%)** | 23.3 | 63.1 | 73.4 | 44.3 | <0.001 |

**Table 2.** Distribution of CDR stages across cohorts.

| Cohort | CDR 0 n (%) | CDR 0.5 n (%) | CDR 1 n (%) | Total |
|---|---|---|---|---|
| **ADNI** | 1538 (36.4%) | 2041 (48.3%) | 644 (15.3%) | 4223 |
| **AIBL** | 1188 (81.3%) | 219 (15.0%) | 55 (3.8%) | 1462 |
| **NIFD** | 275 (94.5%) | 16 (5.5%) | 0 (0.0%) | 291 |
| **Total** | 3001 (50.2%) | 2276 (38.1%) | 699 (11.7%) | 5976 |

### *2.2. Data Preprocessing and Feature Preparation*

T1w imaging was used given its routine clinical availability and suitability for assessing brain atrophy. MRI scans originated from multiple scanners and acquisition protocols without explicit harmonization, reflecting real-world variability and allowing the models to learn representations robust to site and vendor and MRI model-specific differences. Preprocessing included automated brain localization to define the minimal bounding box, center-cropping or zero-padding to a fixed spatial resolution of 140 × 160 × 160 voxels, and voxel intensity normalization to [0,1] per volume to mitigate scanner-dependent variations. Data augmentation was applied during training including random affine transformations, Gaussian noise injection, and spatial blurring to increase robustness and reduce overfitting, with augmentation hyperparameters optimized via validation.

Tabular data comprised demographic and genetic variables with strong relevance to Alzheimer's disease, specifically age, sex, years of education, and APOE genotype encoded as ordinal categorical values to preserve allelic order while maintaining a compact representation. Tabular preprocessing included multivariate imputation via chained equations restricted to the training set to avoid information leakage, standardization of continuous variables, and ordinal encoding of categorical features. This representation allows downstream neural network models to learn both linear and nonlinear interactions among tabular features, which are known to modulate disease onset and progression.

### *2.3. Modeling Approach*

The proposed deep learning framework consists of modality-specific encoders for structural MRI and tabular clinical data, followed by a multimodal fusion module designed to produce clinically meaningful and ordinal-consistent predictions of dementia severity.
MRI volumes were processed using a three-dimensional convolutional neural network based on a 3D ResNet-18 backbone pretrained on large-scale medical imaging datasets [26]. The network extracts hierarchical spatial representations through stacked residual convolutional blocks, enabling the capture of both fine-grained local patterns (e.g., cortical thinning and subcortical structural changes) and global neuroanatomical alterations associated with progressive brain atrophy across CDR stages. Feature maps produced by the final convolutional block were aggregated using adaptive average pooling and projected through fully connected layers to obtain compact latent embeddings suitable for multimodal integration. Batch normalization and residual connections were employed to stabilize training and facilitate gradient propagation in the volumetric architecture.

Tabular inputs, including demographic and genetic variables such as age, sex, APOE genotype, and years of education, were modeled using TABPFN (Tabular Prior-Data Fitted Network), a transformer-based architecture specifically designed for tabular prediction tasks [25]. Unlike conventional multilayer perceptrons, TABPFN leverages a meta-learned prior over tabular datasets, allowing it to capture complex nonlinear relationships and feature interactions without extensive hyperparameter tuning. The model produces latent representations that encode both individual feature contributions and higher-order interactions between demographic and genetic risk factors associated with Alzheimer’s disease progression. These representations were projected into the same embedding space as the MRI features to enable multimodal fusion.

Two strategies for multimodal integration were explored. In the first approach, MRI and tabular embeddings were directly concatenated and passed through fully connected layers to learn joint representations. In the second approach, an attention-based fusion mechanism was implemented to dynamically weight modality-specific features and capture cross-modal dependencies. Self-attention layers refine intra-modality representations, while cross-attention modules allow tabular information to modulate MRI-derived features and vice versa, enabling context-aware integration of imaging biomarkers and clinical risk factors.

To account for the ordered nature of dementia severity, the prediction task was formulated as an ordinal regression problem rather than a conventional nominal classification task. Unlike standard multiclass classification, which treats categories as independent labels, ordinal approaches explicitly model the ranking between disease stages and penalize distant-stage errors more strongly than adjacent misclassifications. In this framework, the multi-class staging problem is decomposed into a sequence of ordered binary decisions corresponding to progressive severity thresholds.

Specifically, two neural ordinal regression formulations were explored: CORAL (Consistent Rank Logits) and CORN (Conditional Ordinal Regression for Neural Networks). CORAL reformulates the ordinal prediction task into $k-1$ binary classifiers representing whether a sample exceeds each severity threshold [18]. These classifiers share a common weight vector while maintaining independent bias terms, enforcing consistent rank ordering across logits and preventing contradictory predictions across thresholds. In contrast, CORN models the ordinal structure through a sequence of conditional binary classifiers, where each classifier estimates the probability of surpassing the next severity level conditioned on the previous thresholds being satisfied. This sequential formulation captures dependencies between neighboring stages and further reduces the likelihood of large ordinal violations. Incorporating these ordinal formulations encourages predictions that better reflect the gradual progression of cognitive impairment represented by the CDR scale.

Training was performed using mini-batch optimization with the Adam optimizer. Hyperparameters, including learning rate, dropout rate, fusion strategy, and network depth, were optimized on the validation set. MRI data augmentation included random rotations, translations, and intensity scaling to improve generalization and reduce overfitting. The explored hyperparameter search space for the unimodal T1w and multimodal Tabular + MRI models is summarized in Table 3.

**Table 3.** Training and hyperparameter search space for T1w and Tab+MRI fusion models.

| Hyperparameter | T1w | Tab+MRIFusion |
|---|---|---|
| Training Configuration<br>Max Epochs | 80 | 70 |
| Early Stopping Patience | 10 | 10 |
| Batch Size | 72 | 72 |
| Optimizer Settings Optimizer | {adam, rmsprop} | {adam, rmsprop} |
| Learning Rate | [10−5,10−3] | $[10^{-5}, 5 \times 10^{-3}]$ |
| LR Pretrained | [10−8,10−6] | [10−8,10−6] |
| L2 Reg (Adam) | {0,0.1,0.01,0.001} | {0,0.1,0.01,0.001} |
| Momentum (RMSprop) | [0.7,0.95] | [0.7,0.95] |
| LR Schedule Factor | [0.01,0.5] | [0.01,0.5] |
| Model Architecture Dropout Probability | {0,0.5} | {0,0.5} |
| Linear Layers | [64] | [[512,64], [256,64]] |

| Fusion Method | – | {linear, attention} |
|---|---|---|
| Loss Configuration Ordinal Method | {CORN, CORAL} | {CORN, CORAL} |
| Data Augmentation<br>Random Blur Probability | [0.1,0.9] | [0.1,0.9] |
| Random Noise Probability | [0.1,0.9] | [0.1,0.9] |
| Random Affine Probability | [0.1,0.9] | [0.1,0.9] |
| Weighted Sampler | {True, False} | True |

Cohort-stratified splits were used during training and evaluation to ensure that model performance reflects cross-cohort generalization rather than dataset-specific bias, which is particularly important given the variability in scanner protocols and clinical assessment practices across the ADNI, AIBL, and NIFD cohorts. The final evaluation was conducted exclusively on this unseen test set. Mean absolute error (MAE) served as the primary metric to assess ordinal staging performance, complemented by adjacent accuracy, quadratic weighted kappa (QWK), and confusion matrices to characterize stage-wise discrimination and misclassification patterns.

## 3. Results

### *3.1 Unimodal Model Performance*

Unimodal experiments were conducted using models trained either on tabular clinical variables or structural T1-weighted MRI data. Performance results are summarized in Table 4, and confusion matrices are shown in Figures 1 and 2.

The tabular model achieved moderate performance with an adjacent accuracy of 0.9588 and a mean absolute error (MAE) of 0.3835, indicating that demographic and genetic variables alone contain relevant information for disease staging but provide limited discriminative power between adjacent CDR levels. Misclassifications were predominantly observed between neighboring stages, particularly between CDR 0.5 and CDR 1. The unimodal MRI model achieved slightly lower overall performance, with an MAE of 0.4794 and a quadratic weighted kappa (QWK) of 0.4435. However, predictions remained largely within adjacent stages, resulting in a high adjacent accuracy of 0.9626, suggesting that structural MRI captures progressive neuroanatomical patterns associated with disease severity despite the increased complexity of volumetric imaging data. Overall, these results demonstrate that both imaging and tabular data provide complementary signals for staging dementia severity, while also highlighting the limitations of unimodal models in capturing the full spectrum of disease variability.

Models were trained and validated using cohort-stratified splits derived from the ADNI, AIBL, and NIFD datasets. A strictly independent held-out test set was constructed by reserving a subset of subjects that were completely excluded from all training, validation, preprocessing (including imputation), and hyperparameter tuning procedures. All splits were performed at the subject level to prevent data leakage. The final evaluation was conducted exclusively on this unseen test set.

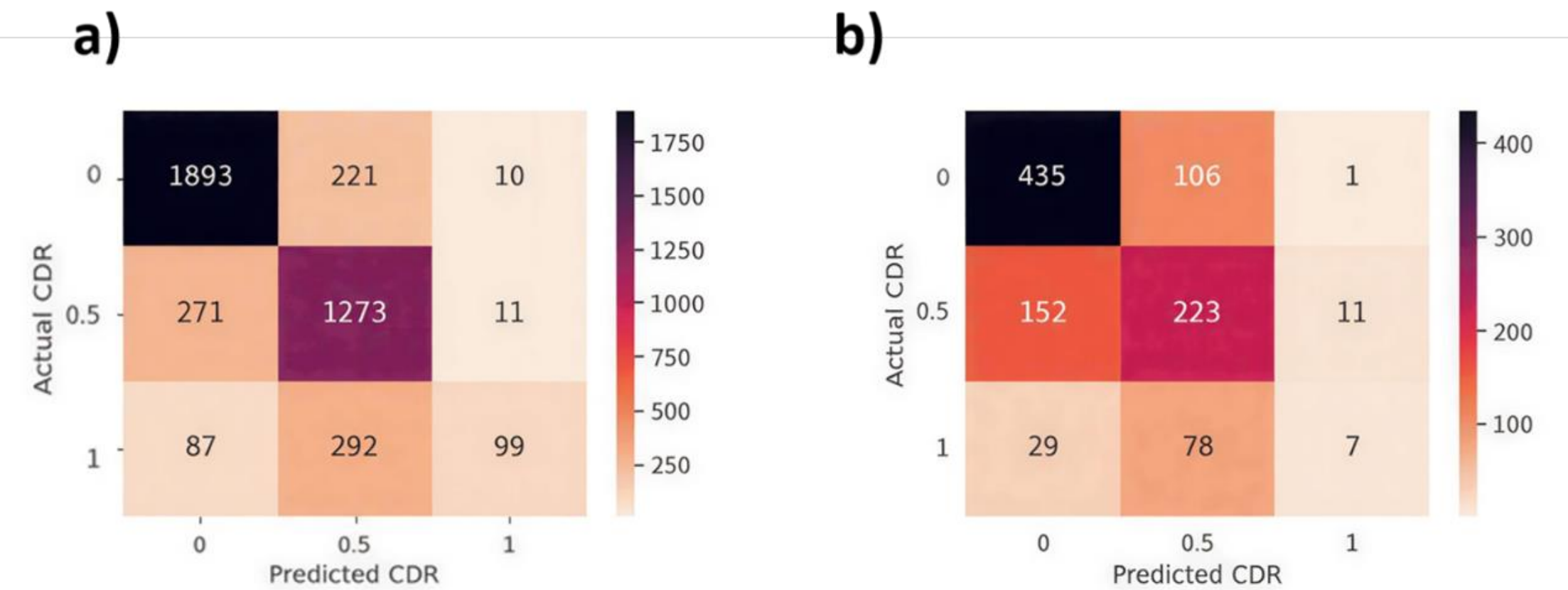

**Figure 1.** Confusion matrices for the unimodal tabular model. (a) Training performance using demographic and genetic variables. (b) Evaluation on the independent test set. Rows denote ground-truth Clinical Dementia Rating (CDR) stages, and columns denote predicted stages. Training confusion matrices reflect the weighted sampling strategy used during training and therefore do not represent the original class distribution.

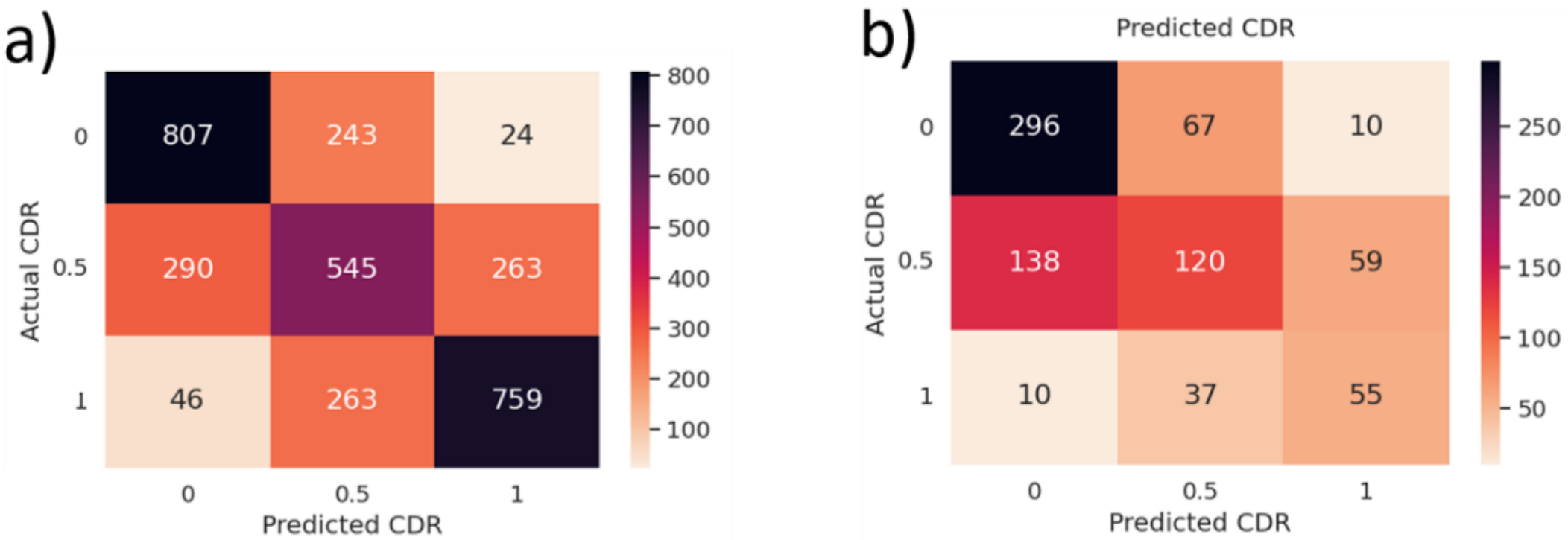

**Figure 2.** Results of the unimodal T1-weighted MRI model using an ordinal CORN regression head. (a) Training confusion matrix. (b) Test confusion matrix. Training confusion matrices reflect the weighted sampling strategy used during training and therefore do not represent the original class distribution.

**Table 4.** Performance comparison across unimodal approaches on train and test. A) Tabular model. B) T1-weighted MRI model.

A)

| Class (CDR) | Precision (train) | Recall (train) | F1-score (train) | Precision (test) | Recall (test) | F1-score (test) |
|---|---|---|---|---|---|---|
| **0** | 0.841 | 0.891 | 0.865 | 0.707 | 0.803 | 0.752 |
| **0.5** | 0.713 | 0.819 | 0.762 | 0.548 | 0.578 | 0.563 |
| **1** | 0.825 | 0.207 | 0.331 | 0.368 | 0.061 | 0.104 |

B)

| Class (CDR) | Precision (train) | Recall (train) | F1-score (train) | Precision (test) | Recall (test) | F1-score (test) |
|---|---|---|---|---|---|---|
| **0** | 0.706 | 0.751 | 0.728 | 0.667 | 0.794 | 0.725 |
| **0.5** | 0.519 | 0.496 | 0.507 | 0.536 | 0.379 | 0.444 |
| **1** | 0.726 | 0.711 | 0.718 | 0.444 | 0.539 | 0.487 |

### *3.2 Multimodal Fusion Improves CDR Staging*

Integrating structural MRI with tabular demographic and genetic information improved staging performance compared to unimodal models. As shown in Table 5, the multimodal ordinal model achieved the highest adjacent accuracy and QWK, while the non-ordinal baseline achieved the lowest MAE. The two modalities capture complementary aspects of disease progression. Structural MRI provides detailed information about neuroanatomical alterations associated with neurodegeneration, while demographic and genetic variables encode established clinical risk factors that are not directly observable in imaging data. By jointly modeling these sources of information, the multimodal approach can better capture the heterogeneous manifestations of cognitive decline.

Multimodal fusion particularly reduced larger staging errors while maintaining high adjacent-stage accuracy, indicating that predictions remain consistent with the gradual progression of the disease. Compared to unimodal approaches, the multimodal model also achieved the highest quadratic weighted kappa (QWK), reflecting improved agreement with the ordered structure of CDR stages and a better alignment with the clinical interpretation of disease severity. These results demonstrate that combining MRI and tabular data provides a more complete representation of disease severity and leads to more accurate and clinically coherent staging predictions.

### *3.3 Effect of Ordinal Modeling*

To assess the impact of explicitly modeling the ordered structure of the CDR scale, the multimodal ordinal model was compared with a non-ordinal baseline trained using standard multiclass classification (see Table 5). The ordinal model achieved a higher quadratic weighted kappa (QWK 0.5492) compared to the non-ordinal model (QWK 0.4772), indicating improved agreement with the ground truth when accounting for the severity distance between categories. Adjacent accuracy was also slightly higher for the ordinal model (0.9704 vs 0.9650), suggesting that predictions produced by the ordinal formulation more frequently fall within clinically neighboring stages. The mean absolute error (MAE) differed between both approaches, with values of 0.3634 for the ordinal model and 0.3400 for the non-ordinal baseline.

Despite this small difference, the higher QWK achieved by the ordinal model suggests that it better preserves the relative ordering of disease severity levels, which is particularly important in clinical scales such as CDR where the progression between categories is gradual rather than categorical. Confusion matrices indicate that misclassifications for both models primarily occurred between adjacent CDR stages (see Figure 3), reflecting the inherent difficulty of distinguishing borderline clinical states. Notably, the ordinal formulation reduces large-severity errors, which are more detrimental in ordinal evaluation metrics and clinical interpretation. Class-specific performance metrics are summarized in Table 6.

**Table 5.** Performance comparison across unimodal and multimodal approaches on an independent test set. For the multimodal approach, the ordinal classification model was compared with a non-ordinal baseline.

| Metric | Tabular | T1-weighted | Tabular + T1-weighted (Non-Ordinal) | Tabular + T1-weighted (Ordinal) |
|---|---|---|---|---|
| **N samples** | 776 | 776 | 776 | 776 |
| **Adjacent Accuracy** | 0.9588 | 0.9626 | 0.9650 | 0.9704 |
| **MAE** | 0.3835 | 0.4794 | 0.3400 | 0.3634 |
| **MSE** | 0.4659 | 0.5541 | 0.4100 | 0.4227 |
| **RMSE** | 0.6826 | 0.7444 | 0.6403 | 0.6501 |
| **QWK** | 0.4329 | 0.4435 | 0.4772 | 0.5492 |
| **Mean Error** | -0.2497 | -0.1598 | -0.2200 | -0.1314 |
| **Std Error** | 0.6353 | 0.7270 | 0.6013 | 0.6367 |

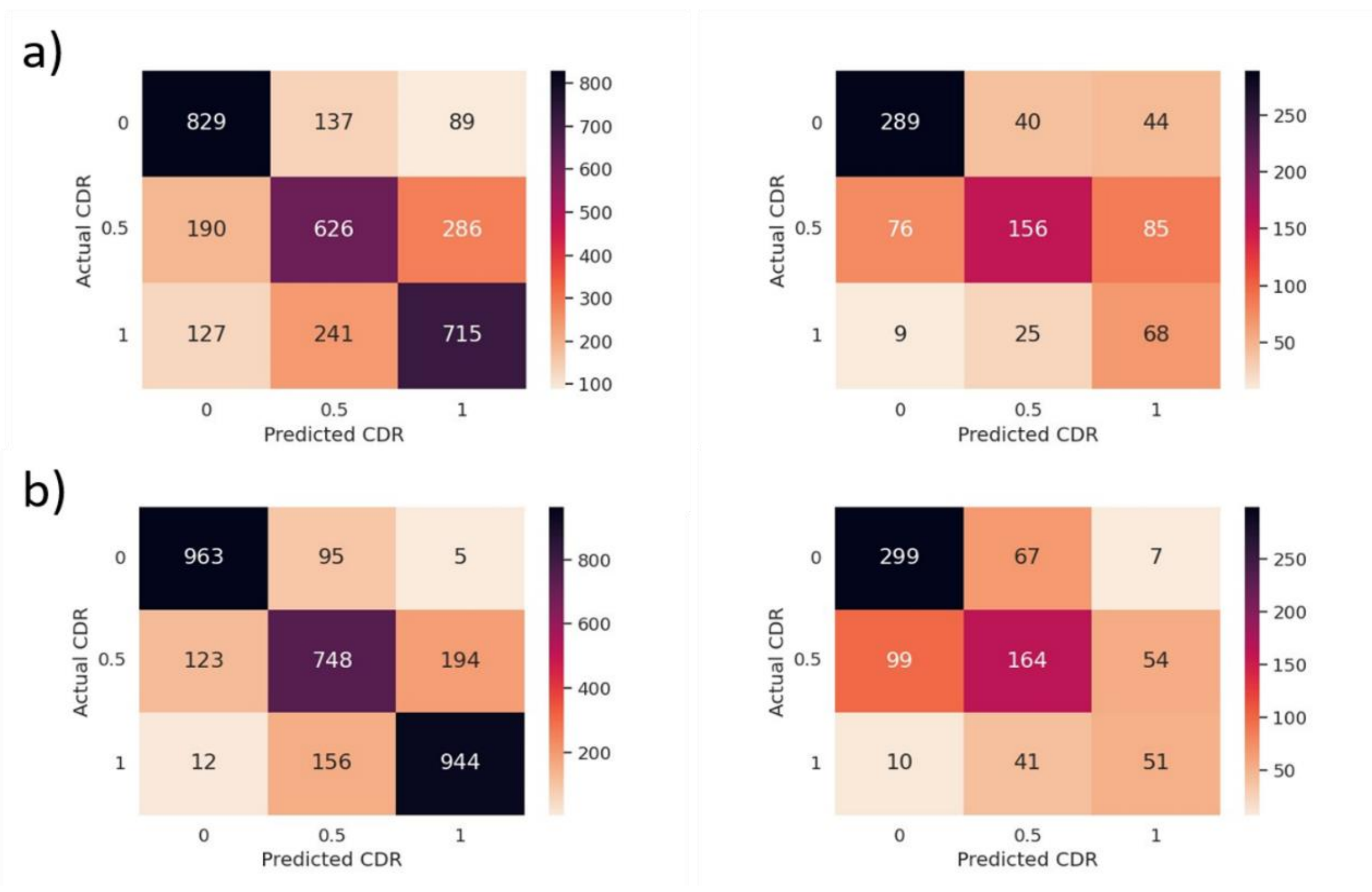

**Figure 3.** Confusion matrices for the training and validation sets. (a) Multimodal fusion model combining T1w MRI images and tabular data, trained *without* incorporating the ordinal component. (b) The same multimodal model trained using the ordinal staging approach based on CORN.

**Table 6.** Performance comparison across multimodal approaches on train and test sets. A) Multimodal model without ordinal staging. B) Our proposed approach by using ordinal staging.

A)

| Class (CDR) | Precision (train) | Recall (train) | F1-score (train) | Precision (test) | Recall (test) | F1-score (test) |
|---|---|---|---|---|---|---|
| **0** | 0.723 | 0.786 | 0.753 | 0.773 | 0.775 | 0.774 |
| **0.5** | 0.624 | 0.568 | 0.594 | 0.706 | 0.492 | 0.580 |
| **1** | 0.656 | 0.660 | 0.658 | 0.345 | 0.667 | 0.455 |

B)

| Class (CDR) | Precision (train) | Recall (train) | F1-score (train) | Precision (test) | Recall (test) | F1-score (test) |
|---|---|---|---|---|---|---|
| **0** | 0.877 | 0.906 | 0.891 | 0.733 | 0.802 | 0.766 |
| **0.5** | 0.749 | 0.702 | 0.725 | 0.603 | 0.517 | 0.557 |
| **1** | 0.826 | 0.849 | 0.837 | 0.455 | 0.500 | 0.477 |

An error analysis revealed that most misclassifications occurred between adjacent CDR stages. In particular, confusion was most frequently observed between CDR 0.5 and CDR 1, reflecting the subtle transition between very mild cognitive impairment and mild dementia. Larger errors spanning multiple disease stages were rare. This pattern is also reflected in the confusion matrices, with predictions concentrated around the diagonal and limited off-diagonal errors.

### *3.4 Robustness Analysis*

To provide a more robust estimate of model performance and account for sampling variability, we computed 95% confidence intervals (CI) using bootstrap resampling on the test set (N = 776). The results are summarized in Table 7. The multimodal model achieved an accuracy of 0.667 (95% CI: 0.631–0.700) and a macro F1-score of 0.591 (95% CI: 0.550–0.629), indicating stable classification performance across resampled test sets. Importantly, the model showed a high adjacent-stage accuracy of 0.970 (95% CI: 0.959–0.982), confirming that most predictions fall within clinically neighboring CDR stages, which is consistent with the ordinal nature of disease progression.

Error-based metrics further support the robustness of the model, with a mean absolute error (MAE) of 0.363 (95% CI: 0.324–0.403) and an RMSE of 0.650 (95% CI: 0.608–0.690). The relatively narrow confidence intervals across all metrics indicate low variability of model performance under resampling, suggesting good generalization to unseen subjects. Finally, the quadratic weighted kappa (QWK) of 0.550 (95% CI: 0.493–0.604) demonstrates moderate-to-strong agreement with clinical staging, while explicitly accounting for the ordinal structure of the CDR scale.

**Table 7**. Performance of the proposed multimodalTABULAR+T1w model with 95% bootstrap confidence intervals (1,000 resampling iterations; N = 776 independent test subjects). Ordinal approach and baseline model.

| Metric | Ordinal model | Non-Ordinal model |
|---|---|---|
| **Accuracy** | 0.6667 [0.6306 – 0.7001] | 0.5817 [0.5418 – 0.6165] |
| **Adjacent Accuracy** | 0.9704 [0.9588 – 0.9820] | 0.9434 [0.9266 – 0.9588] |
| **MAE** | 0.3629 [0.3243 – 0.4028] | 0.4749 [0.4311 – 0.5238] |
| **RMSE** | 0.6497 [0.6077 – 0.6901] | 0.7669 [0.7255 – 0.8149] |
| **QWK** | 0.5497 [0.4926 – 0.6035] | 0.4740 [0.4129 – 0.5281] |
| **F1 (macro)** | 0.5911 [0.5497 – 0.6291] | 0.5100 [0.4742 – 0.5464] |

Thus, these results confirm that the proposed multimodal framework provides not only strong predictive performance but also stable and reliable behavior under uncertainty estimation.

### *3.5 Model Interpretability*

Model interpretability was assessed using Grad-CAM++ for the MRI-based model and SHAP analysis for the tabular component. Grad-CAM++ visualizations highlighted anatomically plausible regions associated with Alzheimer's disease, particularly the hippocampus and surrounding medial temporal lobe structures, which are known to exhibit early neurodegenerative changes. Complementary SHAP analyses identified age and APOE genotype as the most influential tabular predictors contributing to model outputs. Together, these results suggest that the model relies on clinically meaningful imaging and demographic biomarkers when estimating dementia severity, (see Figure 4).

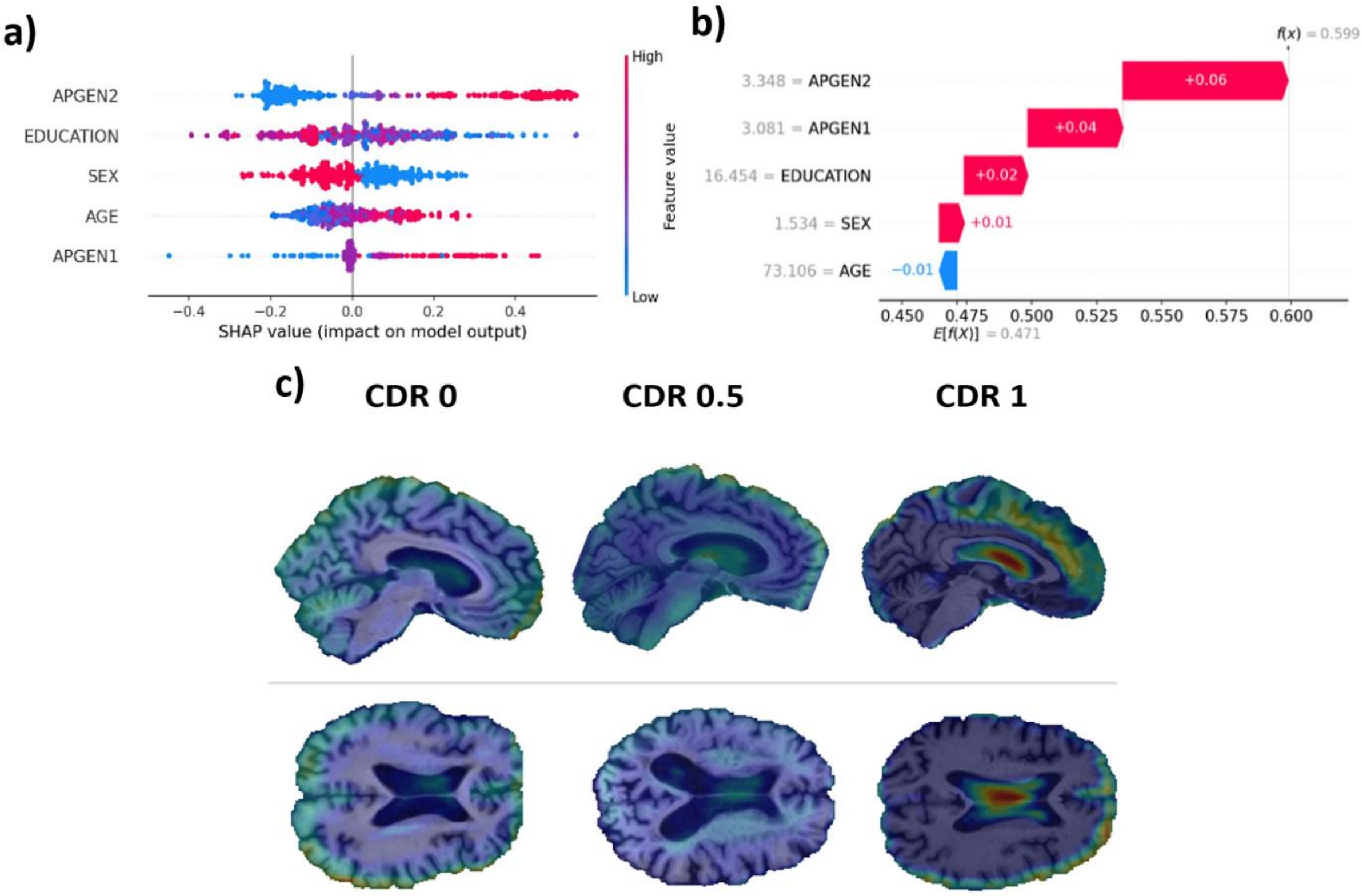

**Figure 4.** SHAP-based interpretability analysis of the tabular model. (a) Global SHAP summary plot showing the contribution of each feature to the model output across all samples. Higher SHAP values indicate stronger influence on predicting higher disease severity. (b) Example SHAP waterfall plot illustrating how individual features contribute to a single prediction. APOE genotype shows the strongest positive contribution, while age and education modulate the prediction in accordance with known clinical risk patterns. (c) Grad-CAM++ visualization maps across all imaging views highlighting the brain regions that most strongly contributed to the model's prediction.

## 4. Discussion

In this work, we introduced an attention-enhanced multimodal deep learning framework for automated staging of Alzheimer's disease according to the Clinical Dementia Rating scale, integrating T1-weighted MRI with demographic and genetic tabular data. By explicitly modeling disease severity as an ordinal variable and combining heterogeneous data sources, the proposed approach addresses key methodological and clinical limitations of prior dementia staging studies, enabling more consistent predictions while maintaining clinical interpretability.

Our results demonstrate that multimodal integration consistently improves staging performance compared with unimodal models. Structural MRI provides substantial discriminative power by capturing characteristic patterns of cortical and subcortical atrophy associated with AD. However, incorporating demographic and genetic variables

- such as age and APOE genotype - further enhances predictive performance. This finding reflects well-established epidemiological evidence that non-imaging risk factors influence disease onset and progression and highlights the practical value of multimodal models that rely on routinely available clinical data rather than specialized biomarkers.
A central component of the proposed framework is the attention-based fusion mechanism, which dynamically weights modality-specific features and captures cross-modal interactions. This capability is particularly important in heterogeneous clinical datasets, where the relative informativeness of imaging and non-imaging modalities may vary across individuals and disease stages. In our experiments, attention-driven fusion improved overall staging performance and reduced large misclassifications, resulting in predictions that more closely align with clinical reasoning. By emphasizing the most informative features for each individual case, the attention mechanism also contributes to improved model interpretability and reliability.

Equally important is the explicit formulation of dementia staging as an ordinal prediction problem. Unlike conventional multi-class classification, ordinal regression encodes the progressive structure of cognitive decline, prioritizing predictions that respect the natural ordering of disease severity. In our evaluation, the multimodal ordinal model achieved the highest agreement with clinical staging and the highest adjacent-stage accuracy, indicating stronger alignment with clinical assessments. While the non-ordinal multimodal model achieved slightly lower absolute prediction error, the ordinal formulation produced predictions that were inherently rank-consistent by design with the ordered structure of the CDR scale and reduced clinically implausible staging errors. These findings suggest that ordinal modeling provides clinically meaningful advantages for staging tasks, even when differences in absolute error metrics are modest. Despite their conceptual suitability, ordinal regression approaches remain relatively underutilized in neuroimaging-based dementia research, and our results provide empirical support for their broader adoption. A methodological consideration is the use of TABPFN for the tabular branch, which does not natively support ordinal regression. TABPFN was selected because of its strong performance on small tabular datasets, whereas ordinal constraints were applied at the multimodal fusion stage where imaging and tabular information were jointly modeled.

Interpretability analyses further reinforce the clinical plausibility of the proposed framework. Saliency maps derived from Grad-CAM++ highlighted medial temporal regions, posterior cingulate cortex, and parietal association areas, key regions known to exhibit early atrophy in Alzheimer's disease and to correlate with disease progression. For tabular inputs, age and APOE ε4 status emerged as the most influential predictors, while education level demonstrated a moderating effect consistent with the concept of cognitive reserve. These findings indicate that model predictions are grounded in biologically and clinically meaningful features rather than spurious correlations, strengthening the interpretability and credibility of the proposed approach.

Models were trained and evaluated using cohort-stratified splits derived from multiple independent research cohorts (ADNI, AIBL, and NIFD), with final performance assessed on a strictly held-out test set consisting of subjects excluded from all training and validation procedures. Although the test data originated from the same source cohorts, strict subject-level separation ensured an unbiased evaluation and prevented data leakage. This strategy provides a more realistic estimate of generalization performance compared with single-cohort evaluations. Despite differences in MRI acquisition protocols, population characteristics, and clinical assessment practices across institutions, the attention-based multimodal models maintained stable performance. These results suggest that training on heterogeneous public cohorts and evaluating on a strictly held-out test set can improve robustness, reduce overfitting, and enhance the potential for real-world clinical applicability.

Nevertheless, several limitations should be acknowledged. Severe dementia stages (CDR 2 and 3) were excluded due to limited sample sizes, restricting the analysis to earlier disease stages. T1-weighted MRI was selected as the primary imaging modality due to its widespread clinical availability, although additional biomarkers such as PET imaging or other MRI modalities such as Fluid-Attenuated Inversion Recovery (FLAIR) or Diffusion-Weighted Imaging may offer additional sensitivity to early pathological changes. In addition, scanner-related variability across datasets was not explicitly harmonized, although the models demonstrated resilience to cross-cohort variability.

Future work could extend this framework to incorporate longitudinal data in order to model disease trajectories and support personalized predictions of cognitive decline. Incorporating uncertainty estimation may further enhance clinical usability by identifying ambiguous cases that require expert review. By relying on routinely acquired MRI and basic clinical variables, the proposed approach provides a scalable, interpretable, and clinically feasible tool for supporting dementia severity assessment. Such systems have the potential to reduce clinical workload, standardize staging practices, and facilitate more consistent and informed decision-making in both research and clinical environments.

## 5. Conclusion and Future Research

This study presents a multimodal deep learning framework for automated dementia staging that integrates T1w MRI with tabular demographic and genetic information. By combining attention-based multimodal fusion with ordinal modeling of CDR stages, the proposed approach explicitly captures the progressive structure of cognitive decline. Experimental results show that multimodal integration improves staging performance compared with unimodal models, while the ordinal formulation yields predictions that better respect the ordered nature of disease severity. In particular, the ordinal models achieved stronger agreement with clinical staging than non-ordinal baselines, indicating that explicitly modeling disease progression can improve the clinical coherence of automated staging systems.

Beyond predictive performance, the framework also provides interpretable insights into the factors driving model decisions. Explainability analyses using Grad-CAM++ and SHAP revealed that predictions rely on anatomically plausible brain regions and clinically relevant variables, supporting transparent and clinically meaningful model behavior.

These findings suggest that combining multimodal learning with ordinal regression can provide a reliable and interpretable foundation for computational dementia staging.

Several directions remain for future research. First, incorporating additional imaging modalities such additional MRI modalities or PET imaging could further improve the characterization of neurodegenerative processes. Second, extending the framework to longitudinal data may allow modeling of disease trajectories and progression dynamics over time. Third, addressing dataset imbalance and improving cross-cohort generalization will be essential for robust deployment in heterogeneous clinical environments. Ultimately, multimodal ordinal learning frameworks hold promise as scalable tools for supporting early detection, monitoring disease progression, and enabling more consistent dementia assessment in both clinical practice and large-scale research studies.

## References

[1] Jong Bin Bae, Subin Lee, Wonmo Jung, Sejin Park, Weonjin Kim, Hyunwoo Oh, Ji Won Han, Grace Eun Kim, Jun Sung Kim, Jae Hyoung Kim, and Ki Woong Kim. Identification of Alzheimer's disease using a convolutional neural network model based on T1-weighted magnetic resonance imaging. *Scientific Reports*, 10(1):22252, December 2020.

[2] Wenzhi Cao, Vahid Mirjalili, and Sebastian Raschka. Rank consistent ordinal regression for neural networks with application to age estimation. *Pattern Recognition Letters*, 140:325–331, December 2020.

[3] Sihong Chen, Kai Ma, and Yefeng Zheng. Med3D: Transfer Learning for 3D Medical Image Analysis, July 2019.

[4] Leonidas Chouliaras and John T. O'Brien. The use of neuroimaging techniques in the early and differential diagnosis of dementia. *Molecular Psychiatry*, 28(10):4084–4097, October 2023.

[5] Eibe Frank and Mark Hall. A Simple Approach to Ordinal Classification. In G. Goos, J. Hartmanis, J. Van Leeuwen, Luc De Raedt, and Peter Flach, editors, *Machine Learning: ECML 2001*, volume 2167, pages 145–156. Springer Berlin Heidelberg, Berlin, Heidelberg, 2001.

[6] Pedro Antonio Gutiérrez, María Pérez-Ortiz, Javier Sánchez-Monedero, Francisco Fernández-Navarro, and César Hervás-Martínez. Ordinal Regression Methods: Survey and Experimental Study. *IEEE Transactions on Knowledge and Data Engineering*, 28(1):127–146, January 2016.

[7] Kaiming He, Xiangyu Zhang, Shaoqing Ren, and Jian Sun. Deep Residual Learning for Image Recognition, December 2015.

[8] Andreas Holzinger, Chris Biemann, Constantinos S. Pattichis, and Douglas B. Kell. What do we need to build explainable AI systems for the medical domain?, December 2017.

[9] Clifford R Jack, David S Knopman, William J Jagust, Leslie M Shaw, Paul S Aisen, Michael W Weiner, Ronald C Petersen, and John Q Trojanowski. Hypothetical model of dynamic biomarkers of the Alzheimer's pathological cascade. *The Lancet Neurology*, 9(1):119–128, January 2010.

[10] Morris Jc. The Clinical Dementia Rating (CDR): Current version and scoring rules. *Neurology*, 43(11), November 1993.
[11] Geert Litjens, Thijs Kooi, Babak Ehteshami Bejnordi, Arnaud Arindra Adiyoso Setio, Francesco Ciompi, Mohsen Ghafoorian, Jeroen A. W. M. van der Laak, Bram van Ginneken, and Clara I. Sánchez. A Survey on Deep Learning in Medical Image Analysis. *Medical Image Analysis*, 42:60–88, December 2017.
[12] Tie-Yan Liu. Learning to Rank for Information Retrieval. *Found. Trends Inf. Retr.*, 3(3):225–331, March 2009.
[13] Scott Lundberg and Su-In Lee. A Unified Approach to Interpreting Model Predictions, November 2017.
[14] Ma Shwe Zin Nyunt, Mei Sian Chong, Wee Shiong Lim, Tih Shih Lee, Philip Yap, and Tze Pin Ng. Reliability and Validity of the Clinical Dementia Rating for Community-Living Elderly Subjects without an Informant. *Dementia and Geriatric Cognitive Disorders Extra*, 3(1):407–416, October 2013.
[15] Sid E. O'Bryant, Laura H. Lacritz, James Hall, Stephen C. Waring, Wenyaw Chan, Zeina G. Khodr, Paul J. Massman, Valerie Hobson, and C. Munro Cullum. Validation of the New Interpretive Guidelines for the Clinical Dementia Rating Scale Sum of Boxes Score in the National Alzheimer's Coordinating Center Database. *Archives of Neurology*, 67(6):746–749, June 2010.
[16] Ramprasaath R. Selvaraju, Michael Cogswell, Abhishek Das, Ramakrishna Vedantam, Devi Parikh, and Dhruv Batra. Grad-CAM: Visual Explanations from Deep Networks via Gradient-based Localization. *International Journal of Computer Vision*, 128(2):336–359, February 2020.
[17] Alberto Serrano-Pozo, Matthew P. Frosch, Eliezer Masliah, and Bradley T. Hyman. Neuropathological Alterations in Alzheimer Disease. *Cold Spring Harbor Perspectives in Medicine*, 1(1):a006189, January 2011.
[18] Xintong Shi, Wenzhi Cao, and Sebastian Raschka. Deep Neural Networks for Rank-Consistent Ordinal Regression Based On Conditional Probabilities. *Pattern Analysis and Applications*, 26(3):941–955, August 2023.
[19] Lee, S. N., Woo, S. H., Lee, E. J., Kim, K. K., & Kim, H. R. (2024). Association between T1w/T2w ratio in white matter and cognitive function in Alzheimer's disease. *Scientific Reports*, 14(1), 7228.
[20] Bilello, M., Doshi, J., Nabavizadeh, S. A., Toledo, J. B., Erus, G., Xie, S. X., ... & Davatzikos, C. (2015). Correlating cognitive decline with white matter lesion and brain atrophy magnetic resonance imaging measurements in Alzheimer's disease. *Journal of Alzheimer's Disease*, 48(4), 987-994.
[21] Van Dyck, C. H., Swanson, C. J., Aisen, P., Bateman, R. J., Chen, C., Gee, M., ... & Iwatsubo, T. (2023). Lecanemab in early Alzheimer's disease. *New England Journal of Medicine*, 388(1), 9-21.
[22] Wyman, B. T., Harvey, D. J., Crawford, K., Bernstein, M. A., Carmichael, O., Cole, P. E., ... & Alzheimer's Disease Neuroimaging Initiative. (2013). Standardization of analysis sets for reporting results from ADNI MRI data. *Alzheimer's & Dementia*, 9(3), 332-337.
[23] Ellis, K. A., Bush, A. I., Darby, D., De Fazio, D., Foster, J., Hudson, P., ... & AIBL Research Group. (2009). The Australian Imaging, Biomarkers and Lifestyle (AIBL) study of aging: methodology and baseline characteristics of 1112 individuals

recruited for a longitudinal study of Alzheimer's disease. *International psychogeriatrics*, 21(4), 672-687.
[24] Raamana, P. R., Rosen, H., Miller, B., Weiner, M. W., Wang, L., & Beg, M. F. (2014). Three-class differential diagnosis among Alzheimer disease, frontotemporal dementia, and controls. *Frontiers in neurology*, 5, 71.
[25] Hollmann, N., Müller, S., Eggensperger, K., & Hutter, F. (2022). Tabpfn: A transformer that solves small tabular classification problems in a second. *arXiv preprint arXiv:2207.01848*.
[26] Alzubaidi, L., Santamaría, J., Manoufali, M., Mohammed, B., Fadhel, M. A., Zhang, J., ... & Duan, Y. (2021). MedNet: pre-trained convolutional neural network model for the medical imaging tasks. *arXiv preprint arXiv:2110.06512*.